\documentclass[letterpaper, 10 pt, journal, twoside]{IEEEtran}
\usepackage[utf8]{inputenc}
\pdfoutput=1

\usepackage[british]{babel}

\usepackage[T1]{fontenc}
\usepackage{graphicx}
\graphicspath{{figures/}}

\usepackage{amsmath,amssymb,amsfonts}

\usepackage[hyphens]{url}
\usepackage[colorlinks=true, allcolors=blue]{hyperref}
\usepackage{balance}
\usepackage{arydshln}

\usepackage[noadjust]{cite}

\newif\iffullekf
\title{4D Radar-Inertial Odometry based on Gaussian Modeling and Multi-Hypothesis Scan Matching}

\author{
Fernando Amodeo$^{1}$, Luis Merino, \textit{Member, IEEE}$^{1}$ and Fernando Caballero$^{1}$%
\thanks{Fernando Amodeo is supported by the predoctoral grant PRE2022-105119 funded by MICIU/AEI/10.13039/501100011033 and by ESF+. %
This work was also partially supported by the following grants: 1) INSERTION PID2021-127648OB-C31, and 2) CO-BUILD PID2024-161069OB-C31 projects, funded by MCIN/AEI/10.13039/501100011033 and by ``\textit{ERDF A way of making Europe}''.}%
\thanks{$^{1}$All authors are members of Service Robotics Laboratory, Universidad Pablo de Olavide (Seville), Spain. {\tt\footnotesize \{famozur, lmercab, fcaballero\}@upo.es}}
\thanks{Copyright 2026 IEEE. Personal use of this material is permitted. Permission from IEEE must be obtained for all other uses, in any current or future media, including reprinting/republishing this material for advertising or promotional purposes, creating new collective works, for resale or redistribution to servers or lists, or reuse of any copyrighted component of this work in other works.}
}

\begin{document}

\maketitle

\begin{abstract}

4D millimeter-wave (mmWave) radars are sensors that provide robustness against adverse weather conditions (rain, snow, fog, etc.), and as such they are increasingly used for odometry and SLAM (Simultaneous Location and Mapping). However, the noisy and sparse nature of the returned scan data proves to be a challenging obstacle for existing registration algorithms, especially those originally intended for more accurate sensors such as LiDAR. Following the success of 3D Gaussian Splatting for vision, in this paper we propose a summarized representation for radar scenes based on global simultaneous optimization of 3D Gaussians as opposed to voxel-based approaches, and leveraging its inherent Probability Density Function (PDF) for registration.
Moreover, we propose optimizing multiple registration hypotheses for better protection against local optima of the PDF.
We evaluate our modeling and registration system against state of the art techniques, finding that our system provides richer models and more accurate registration results.
Finally, we evaluate the effectiveness of our system in a real Radar-Inertial Odometry task.
Experiments using publicly available 4D radar datasets show that our Gaussian approach is comparable to existing registration algorithms, outperforming them in several sequences.

Our code and results can be publicly accessed at:
\texttt{\url{https://github.com/robotics-upo/gaussian-rio-cpp}}
\end{abstract}

\begin{IEEEkeywords}
SLAM; Range Sensing; Sensor Fusion
\end{IEEEkeywords}

\section{Introduction}

\IEEEPARstart{4}{D} millimeter-wave (mmWave) radars have earned considerable popularity in the world of robotics as an alternative to classic camera or LiDAR sensors due to their robustness against adverse weather conditions (rain, snow, fog, etc.), as well as their compact size and low power consumption. As such, there is considerable interest in incorporating these sensors in odometry and SLAM applications. A key component of such system is the modeling and scan matching algorithm. Point clouds are used to capture the geometry of the scene at a given time, and subsequent scans are compared (matched) to the reference model in order to extract localization information. Popular established algorithms that perform this task include the Normal Distributions Transform (NDT) \cite{ndt} or Generalized Iterative Closest Point (GICP) \cite{gicp}. However, despite working well for LiDAR \cite{loam,floam,conlio}, radar point clouds are considerably more challenging to register because they are sparser, noisier, and have a narrower field of view; and, as a result, these algorithms often struggle with the geometry of certain scenes.

We introduce a new method for modeling and registering 4D radar point clouds using 3D Gaussians. Following the success of 3D Gaussian Splatting (3DGS) for vision \cite{3dgs}, we propose the use of a collection of 3D Gaussians for modeling the geometry of a scene.
As in \cite{3dgs} (and in contrast to NDT \cite{ndt}), we simultaneously optimize all parameters of the model, including the position, scaling, rotation and correspondences between points and Gaussians. This approach is entirely featureless, and does not require any priors other than an initial set of centroids.
Moreover, we propose a registration step that allows for multiple simultaneous hypotheses in order to add further robustness against local optimization minima.
We evaluate our Gaussian approach against several established registration algorithms with a test suite independently evaluating the modeling and scan matching, as well as combined within a full Radar-Inertial Odometry (RIO) pipeline using an Extended Kalman Filter (EKF) on multiple datasets and radar types.

This paper is organized as follows: Section~\ref{sec:rw} reviews existing related literature, Sections \ref{sec:gm} and \ref{sec:gsm} describe the proposed Gaussian modeling and scan matching components respectively,
Section~\ref{sec:exp} shows experimental results, and Section~\ref{sec:conc} concludes the paper proposing future work.

\section{Related work} \label{sec:rw}

Registration of incoming point clouds with respect to previous ones is a key technique used in 3D LiDAR and 4D radar odometry and SLAM systems \cite{loam,floam,conlio,_4drtslam,ntu4dradlm,_4dradarslam,efear4d,rivslam,4diriom}.
In particular, the most widely used algorithm for aligning point clouds is Generalized Iterative Closest Point (GICP) \cite{gicp}. This algorithm calculates a distribution probability at each point based on its neighborhood, and optimizes a distribution-to-distribution scan matching function. In the case of radar scans, this is a more difficult task compared to LiDAR because of the lower number and density of points, as well as higher measurement noise/uncertainty in each point's position. For this reason \cite{_4dradarslam} proposed the Adaptive Probability Distribution variant of the algorithm (APDGICP), which leverages GICP's ability to make use of the spatial probability distribution of each point, calculating the required point position covariance matrices according to the radar specifications. However, this is still in essence an approach focused on individual points, without attempting to extract larger scale geometric features that may be more stable and robust for registration purposes.

The Normal Distributions Transform (NDT) algorithm \cite{ndt} subdivides a point cloud using a voxel grid and computes normal distributions for each cell that contains points. Scan matching is performed by maximizing the resulting Probability Density Function (PDF) with respect to an incoming scan. An important limitation of this algorithm is that the grid structure itself imposes restrictions on the geometry, and the resulting model may not be fine grained enough to accurately model the geometry of the environment.
Another problem lies in that discontinuities in the probability function are produced when partially registered points cross cell boundaries during optimization.
NDT has been used with radar scans too \cite{ndtro,ndtro2}.
Voxelized GICP (VGICP) \cite{vgicp} is a derivative that combines NDT's voxel-based representation with the distribution-to-distribution based approach of GICP for registration.
G3Reg \cite{g3reg} is another derivative of NDT, which performs classification on each voxel (plane/cluster/point) in order to merge plane cells together, and extract semantic information to further aid scan matching. In all these algorithms, each individual Gaussian is modeled separately using their associated subcloud extracted from the segmentation. However, we are interested in approaches that jointly optimize all Gaussians, looking at the entire point cloud.

3D Gaussian Splatting (3DGS) \cite{3dgs} is a popular technique for creating a model of a scene from multiple posed images. A 3DGS model is a collection of multiple 3D normal distributions (Gaussians) freely distributed in space, each also containing color/radiance information encoded using spherical harmonics. Even though this technique was originally developed as an alternative to Neural Radiance Fields (NeRF) \cite{nerf} for novel view synthesis, researchers have successfully leveraged this representation to develop SLAM solutions \cite{_3dgsslam1,_3dgsslam2,_3dgsslam3}. Visual-LiDAR methods that improve the robustness of 3DGS with priors sourced from LiDAR input have also been developed \cite{livgs,gslivo,gslivm}. However, 3DGS and these works are all designed for visual input, and they cannot be easily reworked to fully replace it with other sensors such as LiDAR or radar due to the nature of the representation. This affects the rendering process in particular: while NeRF's ray casting framework can be easily adapted to directly predict the output of a Signed Distance Field (SDF) \cite{isdf} (therefore removing the volume rendering component from the system), the differentiable tile rasterizer is a core part of 3DGS that cannot be avoided, neither during training nor inference; thus requiring significant redesign efforts in order to synthesize other types of data such as LiDAR scans \cite{lidargs}.

Following 3DGS as inspiration, we define a purely geometric Gaussian modeling algorithm for radar point clouds based on simultaneous optimization of 3D Gaussians. This representation fundamentally differs from NDT in the fact that Gaussians are not bound to a predefined voxel grid, and all parameters of the model are jointly optimized instead of subdividing the clouds and individually generating a Gaussian for each subcloud.

\section{Gaussian modeling} \label{sec:gm}

Given a point cloud $\mathcal{P}$ containing $M$ points $\{ \mathbf{p}_i | i = 0,1,\ldots,M \}$ with $\mathbf{p}_i \in \mathbb{R}^3$, our goal is creating a summarized representation of its geometry for use with downstream tasks such as mapping or localization.
We model the geometry of a given radar scene using normal distributions, as in \cite{ndt,3dgs,vgicp,g3reg}.

\subsection{Parametrization}

We parametrize the Gaussian model $\mathcal{G}$ of a scene as a collection of $N$ independent trivariate normal distributions (\textbf{3D Gaussians}), each modeling the geometry of an individual region in space. The model is described by the following parameters of each Gaussian $\theta_j = \{ \boldsymbol{\mu}_j, \mathbf{s}_j, \mathbf{q}_j \}$:
\begin{itemize}
\item Center point $\boldsymbol{\mu}_j \in \mathbb{R}^3$, indicating the mean of the distribution.
\item Scaling vector $\mathbf{s}_j \in \mathbb{R}^3$,
indicating the standard deviations in each dimension.
We impose a minimum size restriction so that Gaussians do not become infinitely small (and Mahalanobis distances infinitely large).
\item Rotation quaternion $\mathbf{q}_j \in \mathbb{H}$, compactly encoding
the orientation of the Gaussian.
In order to ensure that the quaternions remain unitary during the optimization, we normalize them prior to usage.
\end{itemize}

Unlike Gaussian Mixture Models (GMMs), we do not define $\pi_j$ weight parameters for the prior probability of a point being generated by a given Gaussian. This is because point cloud data does not encode a complete population of points -- it merely represents a \textit{sample} of said population, biased by factors such as the physical geometric properties of the sensor. For instance, sampling density is inversely proportional to distance from the sensor. This means we cannot model these prior probabilities, and as such we only consider posterior probabilities after a point is assigned to a Gaussian
(using e.g. the shortest Euclidean distance).

The scaling and rotation parameters succinctly describe the covariance matrix of a normal distribution \cite{geocov,forssen04}.
We first derive a rotation matrix $\mathbf{R}_j = \mathbf{R}\{\mathbf{q}_j\}$, and a scaling matrix $\mathbf{S}_j = \mathrm{diag}(\exp_\circ(\mathbf{s}_j))$, where $\exp_\circ(\mathbf{x})$ is element-wise natural exponentiation and $\mathbf{R}\{\mathbf{q}\}$ is
the rotation matrix derived from a given quaternion.
We then define a matrix {$\mathbf{M}_j = \mathbf{R}_j \mathbf{S}_j$} that maps a mean-centered point into the Gaussian's correlation space, and finally derive the covariance matrix {$\boldsymbol{\Sigma}_j = \mathbf{M}_j \mathbf{M}_j^T$} of the distribution.

\subsection{Initialization}

First, it is necessary to determine $N$, that is, the number of 3D Gaussians in the model. We propose choosing $N$ proportionally based on the number of points, so that richer scans can be represented with more Gaussians.
The proportion should also consider the sparsity of the point clouds generated by the radar used.
The model $\mathcal{G}$ is then initialized with all $\mathbf{s}_j$ set to 0 (indicating unit scale), and all $\mathbf{q}_j$ set to the unit quaternion. As for $\boldsymbol{\mu}_j$, we perform Bisecting K-Means clustering to find initial estimates for the centers of the Gaussians. Compared to voxel-based clustering used by NDT or VGICP \cite{ndt,vgicp}, this approach produces fewer spurious Gaussians from outlier points, as K-Means is biased towards higher density areas.

\subsection{Optimization}

$\mathcal{G}$ is optimized iteratively using gradient descent. Each epoch involves the following steps:

\begin{itemize}
\item Each point of $\mathcal{P}$ is matched to the nearest Gaussian center point $\boldsymbol{\mu}_j$. This process builds a set $G_j$ for each Gaussian, containing all matched points. We use the standard Euclidean distance in this step as opposed to the Mahalanobis distance in order to avoid a feedback loop between center and covariance matrix optimization that results in exploding gradients.
\item Each point $\mathbf{p}_i$ is transformed to the coordinate space of its assigned Gaussian as such: $\hat{\mathbf{p}}_i = \mathbf{M}_j^{-1} (\mathbf{p}_i - \boldsymbol{\mu}_j)$. Note that $\mathbf{M}_j^{-1} = \mathbf{S}_j^{-1} \mathbf{R}_j^T$, and $\mathbf{S}_j^{-1}$ is trivial to calculate since $\mathbf{S}_j$ is a diagonal matrix.
\item The parameters of $\mathcal{G}$ (in other words, all $\theta_j$) are updated using the gradient of the loss function.
\end{itemize}

We define the following loss function for each $\theta_j$, derived from the log PDF of the multivariate normal distribution:
\begin{equation}
\mathcal{L}_j = \frac{1}{2|G_j|} \sum_{\mathbf{p}_i \in G_j} \hat{\mathbf{p}}_i^T \hat{\mathbf{p}}_i + \sum_{k}^3 \mathbf{s}_{j_k}.
\end{equation}
The
loss function $\mathcal{L}$ of the entire model $\mathcal{G}$ can be
obtained by
taking the mean of the loss values across all Gaussians.

\section{Gaussian scan matching} \label{sec:gsm}

Given a Gaussian model $\mathcal{G}$ and a point cloud $\mathcal{P}$, we define a registration process that estimates the pose of the robot $\xi$ with respect to $\mathcal{G}$. Moreover, we extend the usual approach to support the simultaneous optimization of multiple hypotheses centered around the current predicted position of the robot, in order to increase the robustness against local optima.
At the end of the optimization, the hypotheses are judged against a scoring function, and the best hypothesis is selected as the output of the scan matching algorithm while the other hypotheses are discarded. Note that this is unlike a classical particle filter, which does persist the hypotheses throughout time.

\subsection{Definitions}

We define the robot pose $\xi$ to be a member of $SE(3)$; that is, it includes a translational component $\mathbf{t} \in \mathbb{R}^3$ and rotational component $\mathbf{q} \in SO(3)$, with respect to \textit{some} frame of reference. We consider that the uncertainty about the true robot pose follows \textit{some} probability distribution: $\xi \thicksim \mathrm{P}$. Given a suitable model of $\mathrm{P}$, we can sample a swarm of $K$ pose particles $\hat{\xi}^k$, each representing a different hypothesis, and use an optimization algorithm that evaluates each particle against $\mathcal{G}$ to further refine these estimates until one of the particles reaches a suitable optimum.

\subsection{Optimization}

The process is performed iteratively using Gauss-Newton. Each iteration involves the following steps:

\begin{itemize}
\item We define $K$ virtual copies of $\mathcal{P}$ called $\mathcal{P}'^k$, each registered according to every pose hypothesis $\hat{\xi}^k$.
\item Every point $\mathbf{p}^k_i$ of $\mathcal{P}'^k$ is matched to the Gaussian $\theta_j$ in $\mathcal{G}$ that results in the lowest Mahalanobis distance, which we call $d^k_i$.
\item We perform the usual Gauss-Newton update step for each particle $\hat{\xi}^k$, using the sums of squared Mahalanobis distances as the function to optimize, as in GICP.
Each point contributes to the gradient according to the following weight:
\begin{equation}
w^k_i = \min \left( 1, \frac{d_\textrm{max}}{d^k_i} \right);
\end{equation}
where $d_\textrm{max}$ is a given Mahalanobis distance threshold.
\end{itemize}

The optimization stops when all particles converge to a solution (i.e. when the update applied is under a given order of magnitude), or when a maximum number of iterations is reached; whichever happens first.
The resulting output pose of the registration algorithm is the particle with the lowest value of the scoring function $\mathcal{L}_k$ defined as follows:
\begin{equation}
\mathcal{L}_k = \frac{1}{M} \sum_{i}^M \min \left(d^k_i, d_\textrm{max} \right),
\end{equation}
where $M$ is the number of points in the scan. We consider a failure condition when the output particle has not converged.

\section{Experimental results} \label{sec:exp}

We use existing public 4D radar datasets for evaluation. In particular, we selected the NTU4DRadLM \cite{ntu4dradlm} and Snail-Radar \cite{snailradar} datasets, because together they cover multiple radar types (Oculii Eagle and ARS548), multiple robot platform types (handcart and car), and adverse conditions. In particular, NTU4DRadLM includes scans captured by an Oculii Eagle radar, and a VectorNav \mbox{VN-100} IMU; while Snail-Radar uses both Oculii Eagle and ARS548 radars together with an Xsens \mbox{MTi-3-DK} IMU, and includes sequences recorded in harsher environment conditions.

We compare our Gaussian approach to three common established scan matching algorithms (NDT, GICP and VGICP).
NDT uses the implementation provided by the well known PCL (Point Cloud Library), while GICP and VGICP use \texttt{small\_gicp} \cite{small_gicp} as the underlying implementation. In the case of Oculii Eagle scans, we set a target of 16 points per Gaussian; while for ARS548 scans, due to the much lower number of returned points, we set a target of 8 points per Gaussian. Moreover, these sparser scans trigger a known issue in PCL's implementation of NDT that cause a program termination, so unfortunately NDT is excluded from experiments involving ARS548.
Regarding our scan matching method, we provide results with 1, 8 and 128 particles wherever applicable.
Other parameters include
$d_\mathrm{max}$ = 4 for scan matching, and 5m/5° for particle dispersion. These parameters were chosen to suit the typical depth range/sparsity relation of a radar.

\subsection{Modeling experiments}

\begin{figure}[t]
\centering
\includegraphics[width=0.48\textwidth]{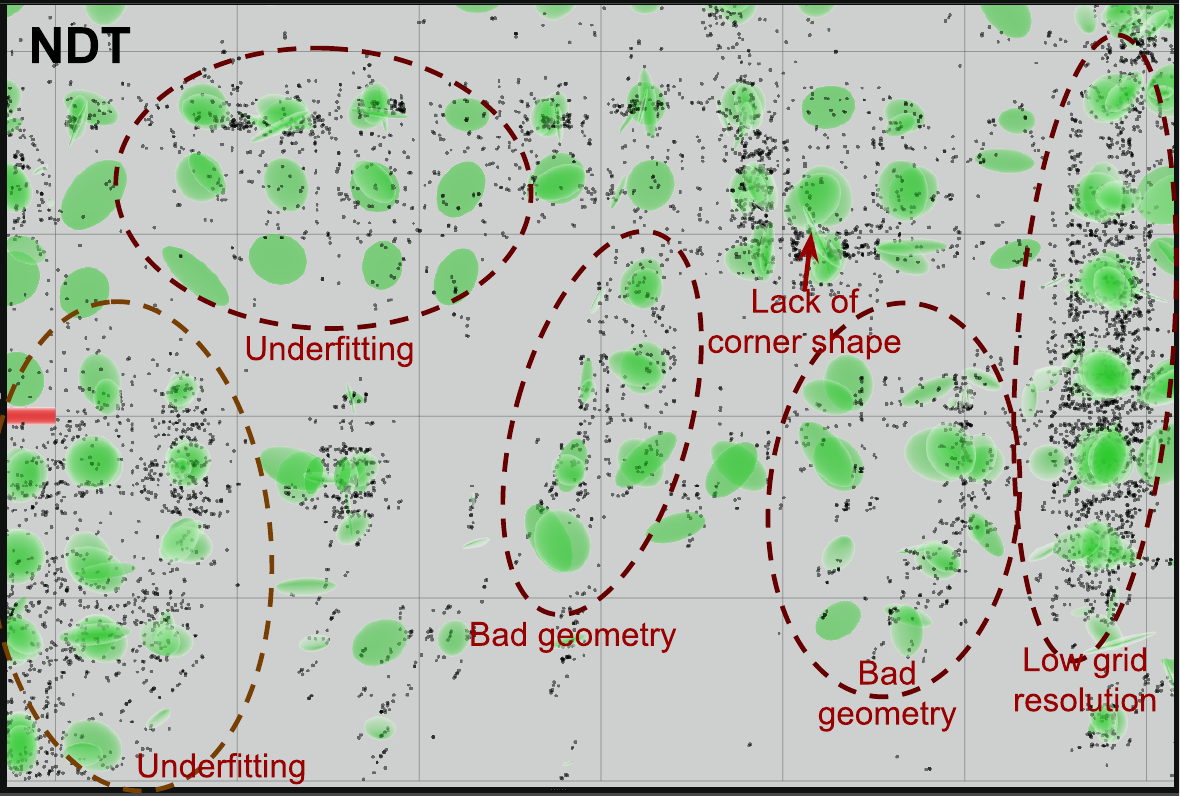}
\vfill
\includegraphics[width=0.48\textwidth]{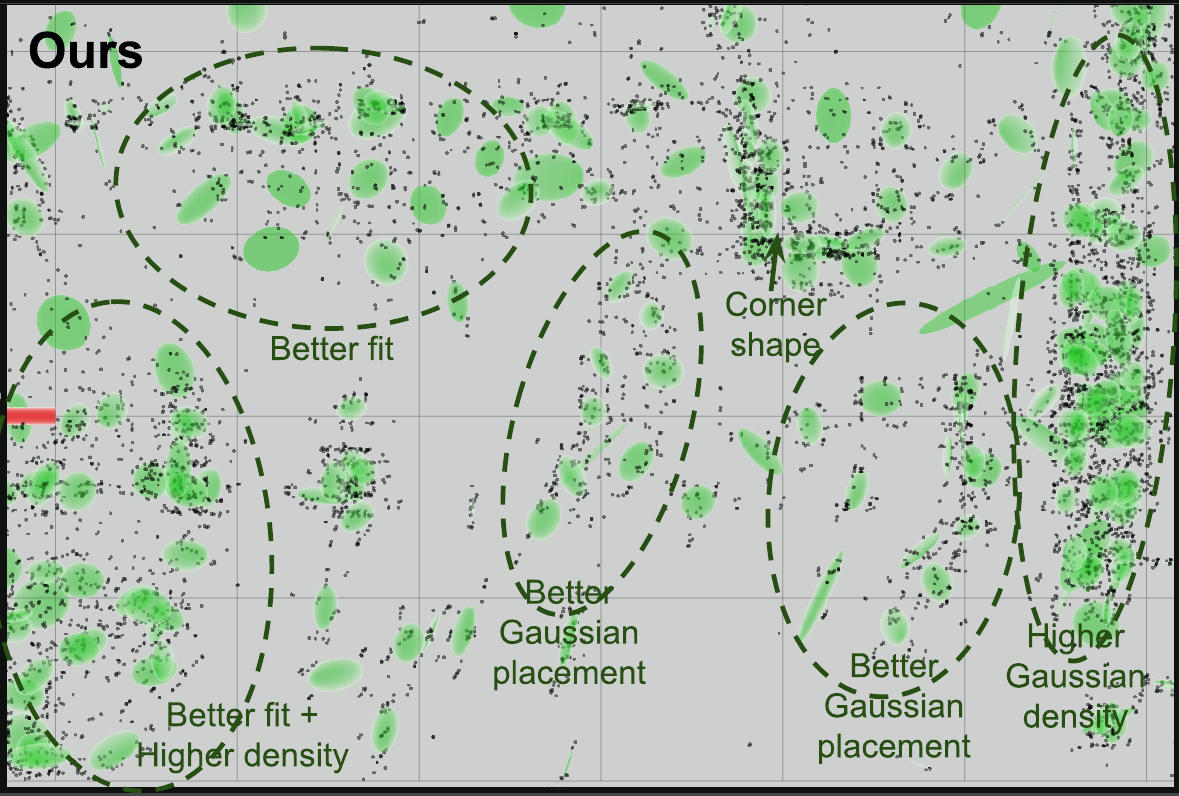}
\caption{Qualitative comparison between NDT and our Gaussian modeling using NTU4DRadLM's cp sequence.}
\label{ndtvsgaussian}
\end{figure}

In order to justify our modeling approach based on globally optimized 3D Gaussians versus existing voxel-based approaches, we compare in Fig.~\ref{ndtvsgaussian} the output of our method against NDT. To do this, we prepare a radar-based map of NTU4DRadLM's \texttt{cp} sequence using the supplied ground truth, and model a section of said map with both NDT and our method. The positions and sizes of the Gaussians generated by NDT are clearly constrained and influenced by the voxel grid, while our approach generates better fitted Gaussians. Moreover, our method is able to preserve geometric details such as corners, and allocates a higher budget of Gaussians to areas with higher point density, resulting in overall better representation of the geometry.

Additionally, we include a video supplement. In the video we show several Gaussian models of the same map created using different points-per-Gaussian ratios. It can be seen that, regardless of said hyperparameter, our modeling method is able to generate a representation reasonably fitting the map. Different number of Gaussians are allocated to areas of the map depending on the density and complexity of the area, as well as the available budget. Moreover, when the ratio is higher (meaning fewer Gaussians), sparse regions with few (outlier) radar points do not receive any Gaussians, which we attribute to our use of K-Means clustering for initialization. Finally, we also include an example of the radar-inertial odometry pipeline on the same NTU4DRadLM \texttt{cp} sequence, compared to the ground truth trajectory.

\begin{figure*}[t]
\centering
\includegraphics[width=\textwidth]{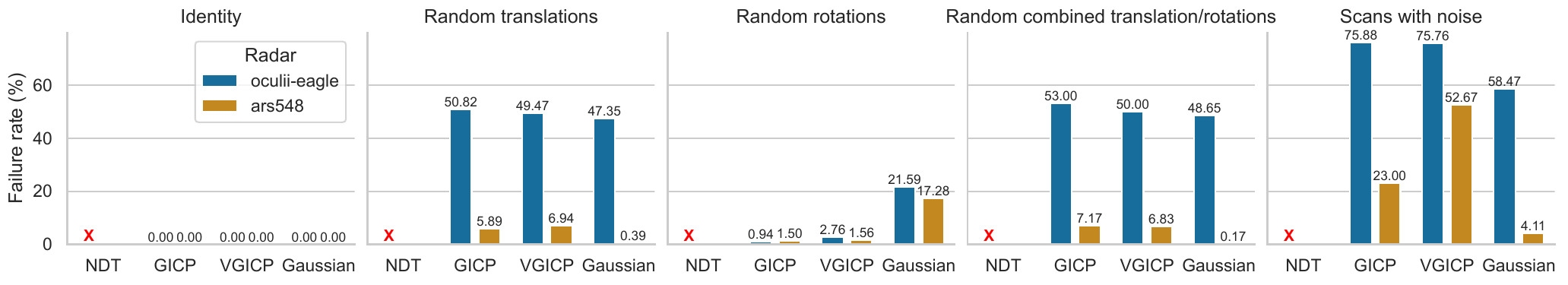}
\vfill
\includegraphics[width=\textwidth]{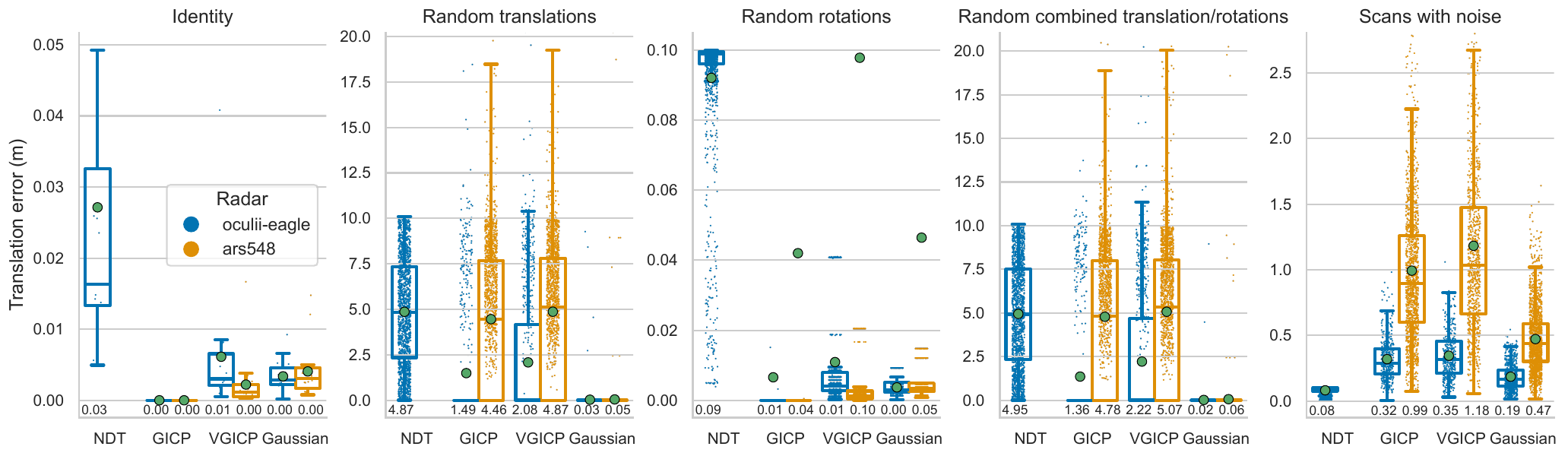}
\vfill
\includegraphics[width=\textwidth]{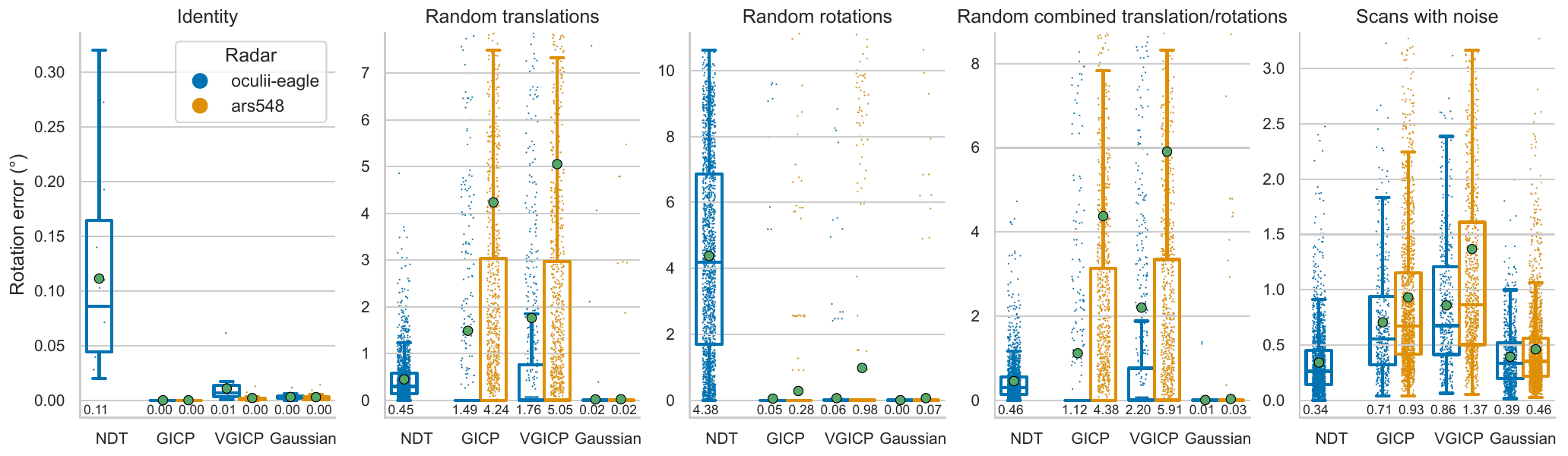}
\caption{Results of the sensibility study, showing failure rate (\%), translation error (m) and rotation error (\textdegree). Lower is better. Green dots represent mean errors, with the numerical value shown below.}
\label{sensigraph}
\end{figure*}

\begin{table*}[ht!]

\caption{Results on NTU4DRadLM Dataset (Oculii Eagle). Best results in \textbf{bold}, second best results \underline{underlined}. Lower is better. Methods marked with \textbf{*} are self-reported metrics. EV: Egovelocity, SM: Scan Matching. Relative translation errors ($t_\mathrm{rel}$) are expressed in \%, relative rotation errors ($r_\mathrm{rel}$) in °/m, and time in ms}
\label{quantit_ntu}
\centering
\begin{tabular}{|c|c c c c|c c|c c|c c|c c|}
\hline
\textbf{Radar/Method} & \multicolumn{3}{|c}{\textbf{Inputs}} & & \multicolumn{2}{|c}{\textbf{cp}} & \multicolumn{2}{|c}{\textbf{nyl}} & \multicolumn{2}{|c}{\textbf{loop2}} & \multicolumn{2}{|c|}{\textbf{loop3}} \\
\hline\hline
\textbf{Oculii Eagle} & IMU & EV & SM & Time & $t_\mathrm{rel}$ & $r_\mathrm{rel}$ & $t_\mathrm{rel}$ & $r_\mathrm{rel}$ & $t_\mathrm{rel}$ & $r_\mathrm{rel}$ & $t_\mathrm{rel}$ & $r_\mathrm{rel}$ \\
\hline
\textit{NTU4DRadLM baseline \cite{ntu4dradlm}: GICP\textbf{*}} & -- & \checkmark & \checkmark & -- & 4.13 & 0.0552 & 4.62 & 0.0184 & 4.84 & 0.0060 & \textbf{3.22} & 0.0060 \\
\textit{NTU4DRadLM baseline \cite{ntu4dradlm}: Fast-LIO\textbf{*}} & \checkmark & -- & \checkmark & -- & 2.94 & 0.0468 & 3.80 & 0.0208 & 7.16 & 0.0057 & 4.55 & 0.0064 \\
\textit{4DRadarSLAM \cite{_4dradarslam}: APDGICP\textbf{*}} & -- & \checkmark & \checkmark & -- & 3.56 & 0.0369 & 3.55 & 0.0171 & 6.09 & 0.0082 & 4.09 & 0.0097 \\
\textit{EFEAR-4D\textbf{*}} \cite{efear4d} & -- & \checkmark & \checkmark & -- & 5.09 & \textbf{0.0125} & 8.93 & 0.0166 & 37.27 & 0.0152 & 37.01 & 0.0175 \\
\textit{RIV-SLAM\textbf{*}} \cite{rivslam} & \checkmark & \checkmark & \checkmark & -- & 2.58 & 0.0342 & --- & --- & 2.69 & 0.0456 & --- & --- \\
EKF-RIO \cite{ekfrio} & \checkmark & \checkmark & -- & -- & 14.76 & 0.3955 & 35.89 & 0.2464 & 44.31 & 0.0349 & 50.23 & 0.0910 \\
\hline
NDT baseline \cite{ndt} & \checkmark & \checkmark & \checkmark & 12.61 & 2.42 & 0.0345 & 7.80 & 0.0408 & 5.61 & 0.0065 & 7.84 & 0.0100 \\
GICP baseline \cite{gicp} & \checkmark & \checkmark & \checkmark & 12.33 & 2.93 & 0.0428 & 2.97 & \textbf{0.0111} & 2.99 & 0.0038 & 3.50 & 0.0042 \\
VGICP baseline \cite{vgicp} & \checkmark & \checkmark & \checkmark & \underline{6.79} & 2.43 & 0.0362 & \underline{2.56} & 0.0136 & 2.73 & \underline{0.0036} & 3.84 & 0.0047 \\
\hdashline
Gaussian (ours) ($K = 1$) & \checkmark & \checkmark & \checkmark & \textbf{5.29} & \underline{2.26} & 0.0339 & \textbf{2.52} & \underline{0.0117} & \textbf{2.26} & \textbf{0.0034} & 3.82 & 0.0043 \\
Gaussian (ours) ($K = 8$) & \checkmark & \checkmark & \checkmark & 19.17 & \textbf{2.06} & \underline{0.0266} & 2.71 & 0.0132 & \underline{2.62} & 0.0039 & \underline{3.41} & \underline{0.0040} \\
Gaussian (ours) ($K = 128$) & \checkmark & \checkmark & \checkmark & 433.52 & 2.31 & 0.0339 & 2.89 & 0.0138 & 2.67 & 0.0042 & 3.43 & \textbf{0.0039} \\
\hline
\end{tabular}
\end{table*}

\begin{table*}[ht!]

\caption{Results on Snail-Radar Dataset (Oculii Eagle and ARS548). Lower is better. Best results in \textbf{bold}, second best results \underline{underlined}. Relative translation errors ($t_\mathrm{rel}$) are expressed in \% and relative rotation errors ($r_\mathrm{rel}$) in mrad/m}
\label{quantit_sr}
\centering
\begin{tabular}{|c|c c|c c|c c|c c|c c|}
\hline
\textbf{Radar/Method} & \multicolumn{2}{|c}{\textbf{st\_20231213\_1}} & \multicolumn{2}{|c}{\textbf{iaf\_20231213\_2}} & \multicolumn{2}{|c}{\textbf{iaf\_20231213\_3}} & \multicolumn{2}{|c}{\textbf{if\_20231213\_4}} & \multicolumn{2}{|c|}{\textbf{if\_20231213\_5}} \\
\hline\hline
\textbf{Oculii Eagle}  & $t_\mathrm{rel}$ & $r_\mathrm{rel}$ & $t_\mathrm{rel}$ & $r_\mathrm{rel}$ & $t_\mathrm{rel}$ & $r_\mathrm{rel}$ & $t_\mathrm{rel}$ & $r_\mathrm{rel}$ & $t_\mathrm{rel}$ & $r_\mathrm{rel}$ \\
\hline
NDT baseline \cite{ndt} & 4.13 & 0.5286 & \textbf{2.04} & \textbf{0.0519} & 3.95 & 0.0653 & 2.86 & \textbf{0.0786} & \textbf{3.15} & \textbf{0.0771} \\
GICP baseline \cite{gicp} & 3.77 & \underline{0.5118} & \underline{2.93} & 0.0629 & 4.05 & 0.0629 & 2.77 & 0.1136 & 4.58 & 0.1572 \\
VGICP baseline \cite{vgicp} & 3.68 & 0.5141 & 3.06 & \underline{0.0610} & 3.17 & 0.0551 & 2.53 & 0.1167 & 4.44 & 0.1489 \\
\hdashline
Gaussian (ours) ($K = 1$) & \underline{3.48} & \textbf{0.5112} & 2.97 & 0.0658 & \underline{2.79} & \underline{0.0537} & \underline{2.48} & 0.1233 & 4.60 & 0.1564 \\
Gaussian (ours) ($K = 8$) & \textbf{3.44} & 0.5165 & 4.15 & 0.0912 & 3.78 & 0.0676 & \textbf{2.33} & \underline{0.0984} & \underline{3.55} & \underline{0.1202} \\
Gaussian (ours) ($K = 128$) & 3.55 & 0.5507 & 3.43 & 0.0833 & \textbf{2.74} & \textbf{0.0480} & 2.34 & 0.1050 & 3.89 & 0.1227 \\
\hline\hline
\textbf{ARS548} & $t_\mathrm{rel}$ & $r_\mathrm{rel}$ & $t_\mathrm{rel}$ & $r_\mathrm{rel}$ & $t_\mathrm{rel}$ & $r_\mathrm{rel}$ & $t_\mathrm{rel}$ & $r_\mathrm{rel}$ & $t_\mathrm{rel}$ & $r_\mathrm{rel}$ \\
\hline
GICP baseline \cite{gicp} & \textbf{1.69} & \underline{0.3049} & \underline{3.36} & \underline{0.0507} & 3.29 & 0.0487 & 2.42 & 0.0761 & \underline{2.26} & \textbf{0.0796} \\
VGICP baseline \cite{vgicp} & 1.95 & \textbf{0.2798} & \textbf{2.57} & \textbf{0.0494} & 3.10 & 0.0491 & 2.81 & 0.0777 & \textbf{2.12} & \underline{0.0836} \\
\hdashline
Gaussian (ours) ($K = 1$) & \underline{1.79} & 0.3561 & 3.37 & 0.0529 & \underline{2.67} & \textbf{0.0380} & 2.37 & \textbf{0.0683} & 3.09 & 0.0902 \\
Gaussian (ours) ($K = 8$) & \underline{1.79} & 0.3561 & 3.56 & 0.0572 & 3.24 & 0.0424 & \textbf{2.14} & 0.0733 & 2.97 & 0.0883 \\
Gaussian (ours) ($K = 128$) & 2.04 & 0.3580 & 3.67 & 0.0608 & \textbf{2.55} & \underline{0.0420} & \underline{2.29} & \underline{0.0697} & 2.61 & 0.0986 \\
\hline
\end{tabular}
\end{table*}

\begin{figure*}[t]
\centering
\includegraphics[width=0.38\textwidth]{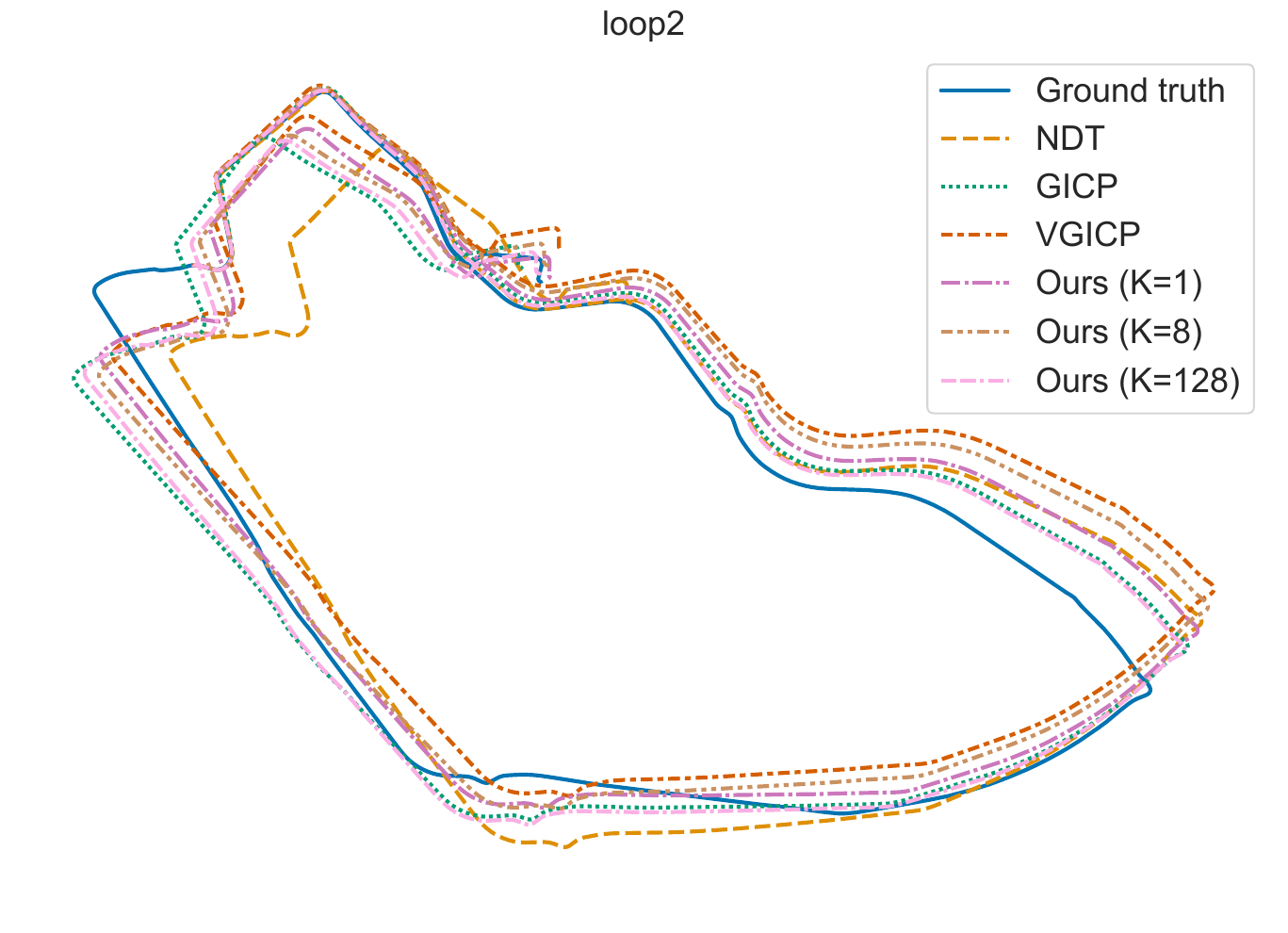}
\includegraphics[width=0.325\textwidth]{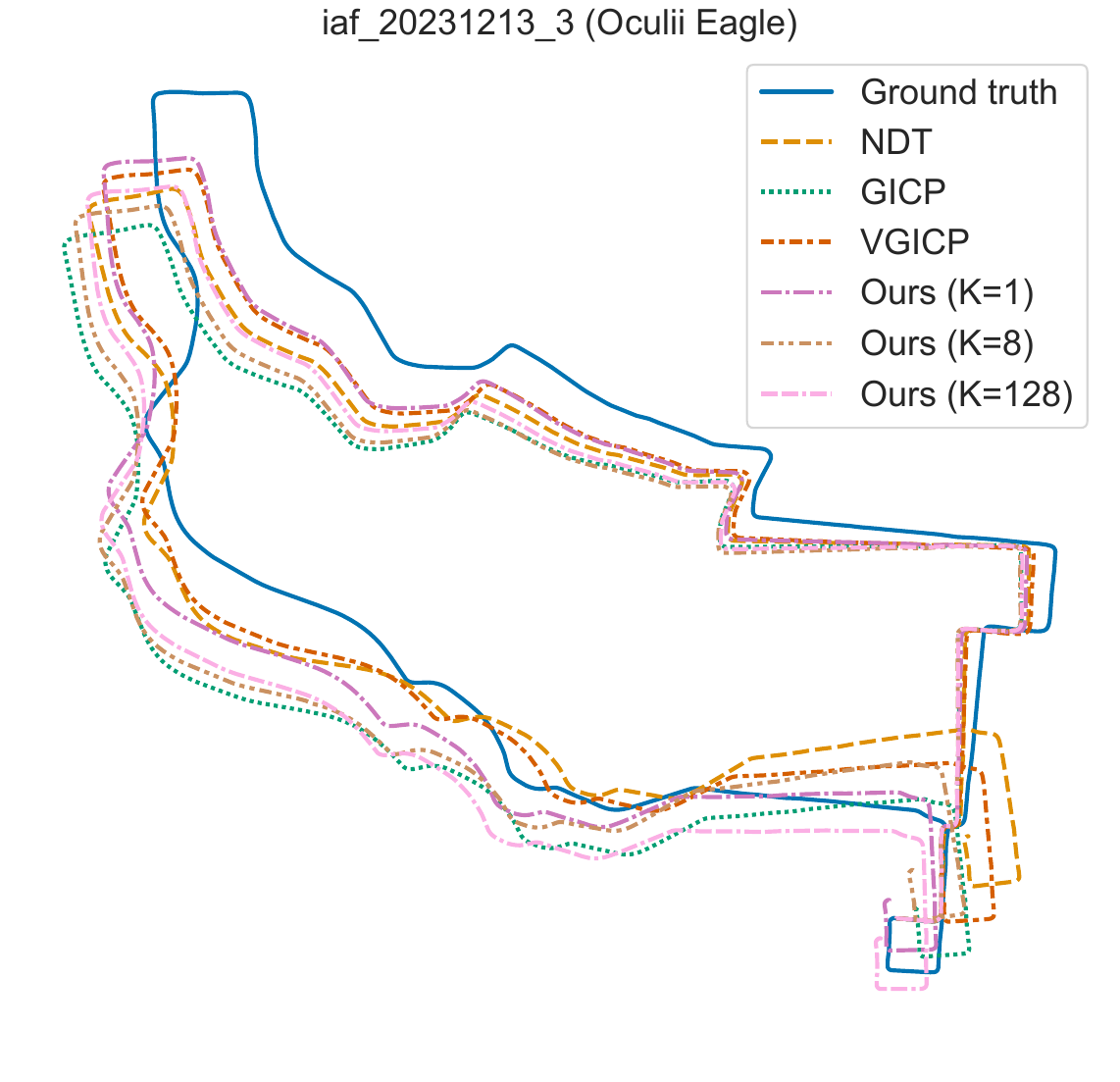}
\includegraphics[width=0.275\textwidth]{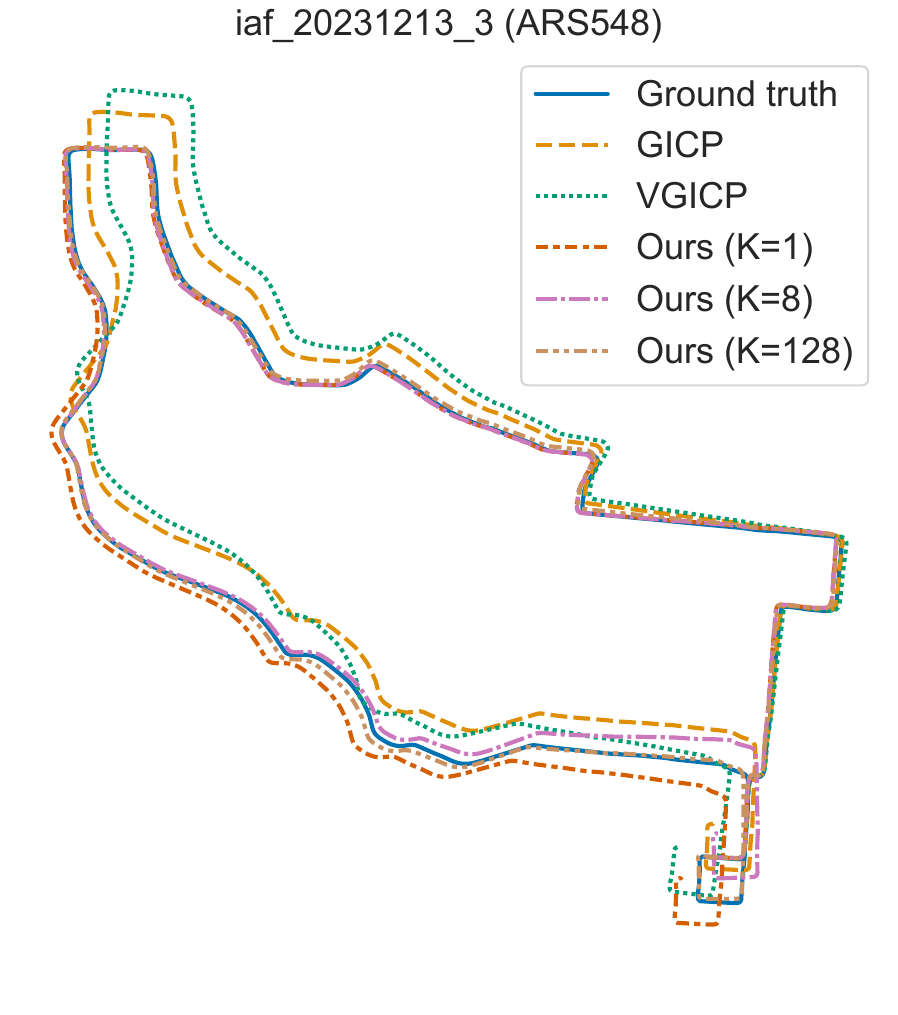}
\caption{Select trajectories of sequences from NTU4DRadLM and Snail-Radar, showing both the ground truth and various different scan matching algorithms used within our odometry pipeline.}
\label{trajplots}
\end{figure*}

\begin{table}[h]
\caption{Results of the ablation study on the modeling approach, replacing global/jointly optimized Gaussian model with local/voxelized VGICP model in our method. Lower is better. Best results in \textbf{bold}. Relative translation errors ($t_\mathrm{rel}$) expressed in \%, relative rotation errors ($r_\mathrm{rel}$) in °/m (NTU4DRadLM) or mrad/m (Snail-Radar).}
\label{ablation}
\centering
\begin{tabular}{|c|c c|c c|}
\hline
\textbf{Radar/Sequence} & \multicolumn{2}{|c}{\textbf{VGICP model}} & \multicolumn{2}{|c|}{\textbf{Gaussian model}} \\
\hline
\multicolumn{5}{c}{} \\
\multicolumn{5}{c}{NTU4DRadLM} \\
\hline
\textbf{Oculii Eagle} & $t_\mathrm{rel}$ & $r_\mathrm{rel}$ & $t_\mathrm{rel}$ & $r_\mathrm{rel}$ \\
\hline
cp    & 4.30 & 0.0651 & \textbf{2.26} & \textbf{0.0339} \\
nyl   & 6.66 & 0.0333 & \textbf{2.52} & \textbf{0.0117} \\
loop2 & 5.81 & 0.0067 & \textbf{2.26} & \textbf{0.0034} \\
loop3 & 6.20 & 0.0081 & \textbf{3.82} & \textbf{0.0043} \\
\hdashline
Avg.  & 5.74 & 0.0283 & \textbf{2.72} & \textbf{0.0133} \\
\hline
\multicolumn{5}{c}{} \\
\multicolumn{5}{c}{Snail-Radar} \\
\hline
\textbf{Oculii Eagle} & $t_\mathrm{rel}$ & $r_\mathrm{rel}$ & $t_\mathrm{rel}$ & $r_\mathrm{rel}$ \\
\hline
st\_20231213\_1  & 3.87 & \textbf{0.4654} & \textbf{3.48} & 0.5112 \\
iaf\_20231213\_2 & \textbf{2.96} & 0.0696 & 2.97 & \textbf{0.0658} \\
iaf\_20231213\_3 & 4.96 & 0.0783 & \textbf{2.79} & \textbf{0.0537} \\
if\_20231213\_4  & 3.23 & \textbf{0.0845} & \textbf{2.48} & 0.1233 \\
if\_20231213\_5  & \textbf{3.57} & \textbf{0.1112} & 4.60 & 0.1564 \\
\hdashline
Avg.             & 3.72 & \textbf{0.1618} & \textbf{3.26} & 0.1821 \\
\hline\hline
\textbf{ARS548} & $t_\mathrm{rel}$ & $r_\mathrm{rel}$ & $t_\mathrm{rel}$ & $r_\mathrm{rel}$ \\
\hline
st\_20231213\_1  & 3.02 & 0.4107 & \textbf{1.79} & \textbf{0.3561} \\
iaf\_20231213\_2 & \textbf{2.62} & \textbf{0.0492} & 3.37 & 0.0529 \\
iaf\_20231213\_3 & 4.07 & 0.0418 & \textbf{2.67} & \textbf{0.0380} \\
if\_20231213\_4  & \textbf{2.02} & 0.0805 & 2.37 & \textbf{0.0683} \\
if\_20231213\_5  & \textbf{1.93} & \textbf{0.0825} & 3.09 & 0.0902 \\
\hdashline
Avg.             & 2.73 & 0.1329 & \textbf{2.66} & \textbf{0.1211} \\
\hline
\end{tabular}
\end{table}

\subsection{Scan matching sensitivity study}

Afterwards, we individually test the stability and sensitivity of the Gaussian scan matching algorithm. For this purpose, we select the \texttt{if\_20231208\_4} sequence from Snail-Radar due to its length (2.23\,km over 8\,min\,35\,s) and its challenging urban geometry. We extract radar scans (from both radars) every 30\,s, which are then modeled with every algorithm (NDT, GICP, VGICP, Gaussians). We perform scan matching experiments between every model and 401 transformed copies of their source radar point clouds: 1 using the identity transformation, 100 using random translations, 100 random rotations, 100 random combined translations/rotations, and 100 copies of the scan with added noise. The random translations and rotations are uniformly sampled between 0 and 10\,m/0\textdegree and 10\textdegree respectively. The direction of translation and axis of rotation is also uniformly sampled. The noise is normally distributed around zero with $\sigma = 1$\,m.

Figure~\ref{sensigraph} shows results for both Oculii Eagle and ARS548 radar scans.
We can observe that our approach produces much lower translation and rotation errors across the board, with a much more compact error distribution. Failure rate is also improved compared to other algorithms in the case of random translations, combined transforms or noise. While it is higher in random isolated rotations, we believe this contributes to rejecting badly converged solutions and thus it can be said that our algorithm behaves more robustly. We must also mention that even though the NDT implementation is capable of reporting convergence failure this never seems to happen (failure rate of 0\%), and thus the errors are much higher and spread out than for the other methods.

We report a single metric for our method in this experiment. This is because nearly all particles converge to the same solution, resulting in our algorithm obtaining the same metrics regardless of particle count. We think this happens because the same scan is used for both modeling and scan matching, and thus the likelihood of converging to a different local minimum is reduced. This also shows a strength of our algorithm: under ``optimal” circumstances the aforementioned near full convergence of all particles occurs even though particle dispersion is high. The next experiment section is dedicated to a full odometry pipeline, where subsequent scans are registered against a model from a previous scan, which allows the multi-hypothesis strategy to take effect.

\subsection{Odometry experiments}

Finally, we perform a comprehensive evaluation on a full Radar-Inertial Odometry task using both datasets and radar types.
For this purpose we develop our own odometry pipeline following well established Extended Kalman Filter practice \cite{ekfstrapdown,ekfrio,4diriom} to tightly integrate all incoming sensor information. Leveraging the Doppler information provided by 4D radar, scans are preprocessed with the RANSAC-LSQ algorithm \cite{ekfrio} to both filter out outlier (dynamic) points, and estimate the current radar-frame velocity of the robot (egovelocity). The filter state includes position, velocity, attitude, accelerometer/gyroscope bias, and radar-to-IMU pose. We perform Kalman propagation using inertial data, and Kalman updates using radar egovelocity observations and scan matching results. In particular for the scan matching we follow \cite{radardofrestriction} and constrain the observation model by only considering 3 degrees of freedom: X/Y/yaw. This is justified in the fact that radars have lower resolution in the Z axis, which degrades the usefulness of Z/roll/pitch information returned by scan matching algorithms.

We evaluate relative translation/rotation errors using the modern and flexible \texttt{evo} evaluation package \cite{evo}, while preserving existing practice by
using a custom evaluation script that follows the behavior of older packages such as \texttt{rpg\_trajectory\_evaluation} as closely as possible. In the case of Snail-Radar sequences, we report rotation errors in mrad/m instead of º/m due to much smaller errors.

The source code of our system, which includes the Gaussian algorithms and the odometry pipeline, can be accessed online on GitHub\footnote{
\url{https://github.com/robotics-upo/gaussian-rio-cpp}
}. We include the script used to generate the odometry trajectories, and the script used to evaluate trajectories with \texttt{evo}. The code also contains all details regarding the system parameters we used, such as EKF covariance initialization, IMU/noise parameters, etc.

\subsubsection{Results on NTU4DRadLM}

We evaluate using the NTU4DRadLM dataset \cite{ntu4dradlm}.
Besides the evaluation performed with our pipeline, we also include previously published results of several radar-focused odometry methods, in order to provide context on the current performance of the state of the art. Due to a lack of Radar-Inertial methods, we also include radar-only methods.
These include the GICP and Fast-LIO baselines tested in the original NTU4DRadLM paper, the APDGICP algorithm tested in the follow-up 4DRadarSLAM paper \cite{_4dradarslam}, as well as other works such as EFEAR-4D \cite{efear4d} or RIV-SLAM \cite{rivslam}.
All works are categorized according to the sensory input used (IMU, egovelocity, scan matching).
We include self-reported metrics as-is, as the trajectory files needed for metric calculation are usually not made public.
The results for EKF-RIO \cite{ekfrio} are generated by running the system directly, as there are no published results on this dataset.
We report metrics without loop closure whenever possible, in order to focus on evaluating pure odometry.

We evaluate the \texttt{cp}, \texttt{nyl} (low speed handcart), \texttt{loop2} and \texttt{loop3} (high speed car) sequences. The \texttt{garden} and \texttt{loop1} sequences are excluded because they contain large gaps in the IMU and radar data, respectively.

Table~\ref{quantit_ntu} shows the comparison between all methods. Our Gaussian method provides the best or second best results for all sequences except for \texttt{loop3}, which seems to favor GICP-based methods (both baseline and our implementation).
Performing multi-hypothesis scan matching helps in some sequences such as \texttt{cp} and \texttt{loop3}. Moreover, a high number of particles is not necessary to obtain these improvements.

Additionally, we report the average scan matching processing time for each method running under our odometry pipeline. The evaluation platform is a desktop PC equipped with an Intel Core i9-9900X CPU and 128 GB of RAM. The single-hypothesis version of our method achieves the fastest scan matching processing times, closely followed by VGICP.
When registering multiple particles our method shows non-linear scaling, as 8 particles only need 3.62x more time than a single particle, while 128 particles need 82x more time.
We also mention that our current RIO implementation is entirely sequential, and that Gaussian modeling at keyframes consumes up to 608.78ms, raising the average scan processing time to 18.13ms. However, we believe that running the modeling and odometry tasks in parallel would remove any timing inconsistencies affecting real-time usage; which we leave as future work.

\subsubsection{Results on Snail-Radar}

In addition to NTU4DRadLM, we perform additional testing on the Snail-Radar dataset \cite{snailradar}. In particular, we selected the five sequences recorded on 2023/12/13 because of the nighttime and rain conditions of that day under a high speed car platform. The longest tested sequence (iaf\_20231213\_2) is 15 min in length, while the others are around 7 min.

Table~\ref{quantit_sr} shows results using the Oculii Eagle radar. In general, we can observe the best results to be spread across the different methods.
Nonetheless, our Gaussian method obtains the best or second best results in 8 out of 10 metrics, and the best translation error in 3 out of 5 sequences (the other two favoring NDT).
Our Gaussian approach produces the best translation errors in 3 out of 5 sequences, and the best rotation errors in 2 out of 5 sequences.
The multi-hypothesis approach is capable of improving performance in 4 out of 5 sequences. Regarding the number of particles, a higher number benefits 3 out of 5 sequences, which indicates much more challenging geometry compared to NTU4DRadLM.
We also note that NDT seems to behave much better in Snail-Radar than in NTU4DRadLM.

Table~\ref{quantit_sr} also shows results using the ARS548 radar. This radar produces much sparser point clouds than Oculii Eagle (around 10x fewer points).
We believe this to be the effect of Oculii Eagle's scan enhancement algorithm.
However, we can observe better metrics in general compared to Oculii Eagle.
The results once again show different methods leading in different sequences,
however it is worth pointing out that our method produces the best results in 2 out of 5 sequences.

\subsubsection{Ablation study}

In order to show the individual contribution of our global, jointly optimized Gaussian modeling approach, we perform an ablation study based on the odometry experiment combining our scan matching implementation with a strong locally constructed/voxelized baseline for modeling (VGICP). That is, we use the Gaussian models generated by VGICP with our scan matching approach and a single particle. Table~\ref{ablation} shows the results on all tested sequences. In the case of NTU4DRadLM our global models clearly obtain the best metrics on all sequences. For Snail-Radar the improvements are more marginal and reduced to some sequences, however our approach still obtains the best translation errors on average with both radars, while obtaining the best rotation errors on average with ARS548.

\subsubsection{Qualitative results}

Fig.~\ref{trajplots} shows plots of select sequences from both datasets. Specifically, we plot the \texttt{loop2} sequence from NTU4DRadLM and the \texttt{iaf\_20231213\_3} sequence from Snail-Radar, the latter for both Oculii Eagle and ARS548 radars. These two sequences are especially long and complex, which allows for qualitatively evaluating the performance of the different methods. We can observe that NDT drifts the most in \texttt{loop2}, while the rest stay fairly close to each other, and in particular our algorithm with 128 multi-hypothesis particles is the closest to the ground truth, closely followed by the 8 particle and 1 particle (single hypothesis) versions. As for \texttt{iaf\_20231213\_3}, the algorithms generally perform better with the ARS548. The 128 particle version of our algorithm is especially close to the ground truth using both radars.

\subsection{Summary of results}

We consider that the results above validate our methodology. Our Gaussian modeling and scan matching algorithm produces comparable or superior relative translation and rotation metrics to existing registration algorithms, as shown by the experiments. Moreover, in several sequences across several datasets and radar types our algorithm produces the best overall results.
On the other hand, although the multi-hypothesis approach provides improvements in some sequences, it also produces regressions in others, indicating the need for further adjustments.

As an additional observation, it is worth mentioning that we believe the radar-to-IMU calibration file provided by NTU4DRadLM to be inaccurate, as it does not seem to match the placement of the different sensors in the photo they provide of their platform. The effects of this inaccurate calibration can be especially seen in the rotational errors, which are much higher for all methods compared to those achieved under Snail-Radar sequences.

\section{Conclusion} \label{sec:conc}

In this paper we introduce a new modeling approach for radar point clouds based on globally and jointly optimized 3D Gaussians unlike existing voxelized or locally constructed approaches \cite{ndt,gicp,vgicp}. We show how this representation produces more robust and better performing scan matching, which we also propose to perform using multiple registration hypotheses.
Through odometry experiments in multiple datasets with multiple radar types, we show that our approach produces results comparable to existing registration algorithms, managing to outperform them in several sequences. At the same time, our method offers higher expandability and faster scan matching compared to existing locally constructed or voxelized approaches \cite{gicp,ndt,vgicp,g3reg}.

Future works include improving the run-time efficiency of the Gaussian modeling (replacing gradient descent, and running in parallel to the odometry pipeline), dynamic management of Gaussians (as in 3DGS), as well as incorporating radar-specific information such as RCS (Radar Cross Section) values, and improving the formulation of the multi-hypothesis particle system with kinematic priors.

% To balance between two columns the last page
\balance

\bibliographystyle{IEEEtran}
\bibliography{IEEEabrv,allthecites}

\end{document}